\documentclass[final,twoside]{IEEEtran}
\usepackage{} 

\usepackage{ulem}
\usepackage{float}
\usepackage{color}
\usepackage{amssymb,amsmath,amsthm,bm}

\makeatletter
\newcommand{\tpmod}[1]{{\@displayfalse\pmod{#1}}}
\makeatother

\usepackage{graphicx,color}

\graphicspath{{figures-pdf/}}
\usepackage{cite}
\usepackage{url}
\usepackage{diagbox}
\usepackage[aboveskip=1pt]{subcaption}

\usepackage{algorithm}
\usepackage{bbding}
\usepackage{pifont}
\usepackage{algpseudocode}
\usepackage{multirow,bigstrut}
\usepackage{tabularx}
\usepackage{arydshln}
\usepackage{empheq}
\usepackage{datetime}
\usepackage{algorithmicx}
\usepackage{algpseudocode}
\usepackage{amsmath}

\newlength\imagewidth
\newlength\figsep
\usepackage[bookmarks=false]{hyperref}
\hypersetup{
 linktocpage=true, pdfborderstyle={/S/S/W 1}, hyperindex=true, bookmarks=true, bookmarksopen=true, bookmarksnumbered=true,
}
\hypersetup{hidelinks}

\usepackage{etoolbox}

\usepackage{fancyhdr}
\begin{document}

\title{3DSS-Mamba: 3D-Spectral-Spatial Mamba for Hyperspectral Image Classification}

\author{Yan~He,~\IEEEmembership{Student Member,~IEEE,}
	Bing Tu,~\IEEEmembership{Senior Member,~IEEE,}
    Bo Liu,~\IEEEmembership{Member,~IEEE,}
	Jun Li,~\IEEEmembership{Fellow,~IEEE,} 
    and Antonio Plaza,~\IEEEmembership{Fellow,~IEEE}
    \thanks{This work was supported in part by the National Natural Science Foundation of China under Grant 62271200; in part by the Startup Foundation for Introducing Talent of Nanjing University of Information Science and Technology (NUIST) under Grant 2023r091; (Corresponding author: Bing Tu.)}
	\thanks{Yan He, Bing Tu and Bo Liu are with the Institute of Optics and Electronics, the Jiangsu Key Laboratory for Optoelectronic Detection of Atmosphere and Ocean, and the Jiangsu International Joint Laboratory on Meteorological Photonics and Optoelectronic Detection, Nanjing University of Information Science and Technology, Nanjing, Jiangsu 210044, China (e-mail: 975861884@qq.com; tubing@nuist.edu.cn; bo@nuist.edu.cn.}
    \thanks{Jun Li is with the Faculty of Computer Science, China University of Geosciences, Wuhan 430074, China (e-mail: lijuncug@cug.edu.cn).}
    \thanks{Antonio Plaza is with the Hyperspectral Computing Laboratory, Department of Technology of Computers and Communications, Escuela Politecnica, University of Extremadura, 10003 C¨¢ceres, Spain (e-mail: aplaza@unex.es).}
}


\markboth{}{}


\maketitle

\begin{abstract}
Hyperspectral image (HSI) classification constitutes the fundamental research in remote sensing fields. Convolutional Neural Networks (CNNs) and Transformers have demonstrated impressive capability in capturing spectral-spatial contextual dependencies. However, these architectures suffer from limited receptive fields and quadratic computational complexity, respectively. Fortunately, recent Mamba architectures built upon the State Space Model integrate the advantages of long-range sequence modeling and linear computational efficiency, exhibiting substantial potential in low-dimensional scenarios. Motivated by this, we propose a novel 3D-Spectral-Spatial Mamba (3DSS-Mamba) framework for HSI classification, allowing for global spectral-spatial relationship modeling with greater computational efficiency. Technically, a spectral-spatial token generation (SSTG) module is designed to convert the HSI cube into a set of 3D spectral-spatial tokens. To overcome the limitations of traditional Mamba, which is confined to modeling causal sequences and inadaptable to high-dimensional scenarios, a 3D-Spectral-Spatial Selective Scanning (3DSS) mechanism is introduced, which performs pixel-wise selective scanning on 3D hyperspectral tokens along the spectral and spatial dimensions.  Five scanning routes are constructed to investigate the impact of dimension prioritization. The 3DSS scanning mechanism combined with conventional mapping operations forms the 3D-spectral-spatial mamba block (3DMB), enabling the extraction of global spectral-spatial semantic representations. Experimental results and analysis demonstrate that the proposed method outperforms the state-of-the-art methods on HSI classification benchmarks.
\end{abstract}

%
\IEEEpeerreviewmaketitle

\section{Introduction}
\label{sec:intro}
\IEEEPARstart{H}{yperspectral} images (HSI) are represented by hundreds of continuous spectral bands in the electromagnetic spectrum, encompassing abundant spectral and spatial information. Compared with natural images, HSI performs widespread applications in various remote sensing scenarios, such as mineral exploration \cite{ni2020mineral, siebels2020estimation}, military reconnaissance \cite{ardouin2007demonstration, peyghambari2021hyperspectral}, and environmental monitoring \cite{camps2013advances}. As a fundamental task for HSI processing, HSI classification focuses on pixel-level category distinguishing for ground objects, which has received considerable attention in remote sensing \cite{ahmad2020fast, 8697135}.

Traditional research on HSI classification typically draws upon spectral feature extraction with hand-crafted descriptors and subspace learning, such as support vector machine (SVM) \cite{melgani2004classification}, linear discriminant analysis (LDA) \cite{camps2013advances}, and manifold learning \cite{huang2019dimensionality, lunga2013manifold}. To cope with the challenges of spectral variability and spectral confusion, several efforts integrate complementary spatial contextual information with spectral features for precise HSI classification mapping, including extended morphological profiles (EMP) \cite{fauvel2008spectral}, extended multi-attribute profiles (EMAP) \cite{dalla2010extended}, and sparse manifold representations \cite{duan2021semisupervised}. However, these methods heavily rely on prior parameter settings, which exhibit insufficient data fitting and description capabilities when confronting complex environmental conditions.

The rapid development of deep learning technology has brought significant paradigms for HSI classification task. Representative models encompass autoencoders (AEs) \cite{10103688, mei2019unsupervised}, convolutional neural networks (CNNs) \cite{sun2019spectral, zhu2020residual, li2019deep, yue2024s2tnet}, recurrent neural networks (RNNs) \cite{8960629, liu2017bidirectional}, and graph convolutional networks (GCNs) \cite{liao2023class, 10226236}. Building upon the properties of local receptive fields and parameter sharing, CNN architectures progressively demonstrate predominant status in HSI classification. For instance, Hu et al. \cite{hu2015deep} first presented a hierarchical 1-D CNN network to extract the high-level spectral features along the spectral dimension of hyperspectral data. Given the characteristics of abundant spectral channels and strong spatial correlation in HSIs, Yang et al. \cite{yang2017learning} constructed a dual-branch architecture that combines 1-D CNN and 2-D CNN to simultaneously capture finer spectral and spatial features for HSI classification. Compared with the 2-D convolutional paradigm restricted to spatial dimension, the 3-D convolutional kernels enjoy the advantage of spectral-spatial joint feature extraction. Classically, Zhong et al. \cite{zhong2017spectral} developed a 3-D CNN-based spectral-spatial residual network (SSRN), which is capable of capturing deep spectral-spatial blocks directly from raw 3-D HSI cubes without additional feature engineering. Despite achieving encouraging performance compared to traditional approaches, CNN-based models struggle to establish long-range dependencies between pixels, failing to capture global spectral and spatial characteristics.
\begin{figure}[tb]
    \centering
    \includegraphics[width=0.45\textwidth]{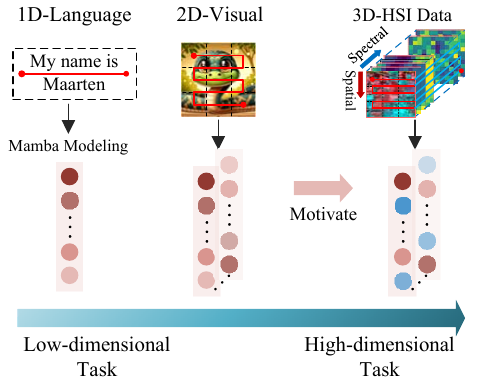}
    \caption{The motivation of the proposed 3DSS-Mamba. Mamba modeling has demonstrated substantial potential in low-dimensional scenarios such as 1D language and 2D visual tasks, motivating its adaptability to high-dimensional HSI classification task.}
    \label{fig:Introduction}
\end{figure}

Benefiting from the powerful long-distance sequence modeling capability based on attention mechanism, Transformer architecture has been adeptly investigated for HSI classification task \cite{yang2022hyperspectral, roy2023spectral, zhang2022convolution, 10038735}. The Vision Transformer (ViT) \cite{dosovitskiy2020image} treats each pixel within the HSI cube as a sequence input to the standard Transformer model, capturing the correlations between pixels through the self-attention mechanism. Derived from this, He et al. \cite{he2019hsi} proposed a bidirectional encoder representation transformer network (HSI-BERT) for HSI classification, which overcomes the restrictions of spatial distance through global receptive fields. Given the significance of long-range dependencies in spectral dimensions, Hong et al. \cite{hong2022spectralformer} devised a novel Transformer-based SpectralFormer (SF) network, which constructs group-wise spectral embeddings to capture the spectral sequence information between neighboring HSI bands. To comprehensively integrate both spectral and spatial information for HSI classification, Zhong et al. \cite{zhong2022spectral} developed a spectral-spatial transformer network (SSTN), which breaks the long-range limitations by integrating spatial attention and spectral association modules, and incorporates a factorized architecture search model to determine the layer-level operations and block-level orders. Additionally, Peng et al. \cite{peng2022spatial} constructed a dual-branch spatial-spectral transformer with cross-attention (CASST), where the spectral branch establishes dependencies among spectral sequences and the spatial branch captures fine-grained spatial contexts. The interaction between spatial and spectral information is performed within each transformer block through a cross-attention mechanism. Although Transformer architecture has exhibited impressive capability in HSI classification, its inherent self-attention mechanism is characterized by quadratic computation complexity $\mathcal{O}\left(N^{2}\right)$, which poses significant challenges in modeling efficiency and memory overhead.

Comparatively, recent State Space Models (SSMs) establish long-distance dependency through state transitions, enjoying the promising attributes of linear computational complexity and scalability. As an effective alternative to the Transformer, Mamba \cite{gu2023mamba} introduces the selective SSMs for 1-D sequence modeling along specific orientation, which demonstrates substantial potential in natural language processing (NLP) tasks \cite{mehta2023long}. To accommodate vision scenarios involving 2D-spatial awareness, Vim \cite{zhu2024vision} and VMamba \cite{liu2024vmamba} extend the Mamba architecture by introducing a multi-directional scanning mechanism to achieve global contextual modeling, showcasing great efficiency and effectiveness in 2D visual tasks, such as object detection and semantic segmentation. Although Mamba architectures have demonstrated substantial potential in low-dimensional scenarios, the adaptability to high-dimensional HSI classification tasks involving 3D hyperspectral data requires further exploration, as depicted in Fig.~\ref{fig:Introduction}.

To this end, this work investigates 3D-Spectral-Spatial Mamba (3DSS-Mamba), an efficient global spectral-spatial contextual modeling framework based on the State Space Model for HSI classification. The 3DSS-Mamba consists of a Spectral-Spatial Token Generation module (SSTG), multiple stacked 3D-Spectral-Spatial Mamba Blocks (3DMB), and a prediction module. Specifically, SSTG converts the HSI cube into a set of spectral-spatial tokens by introducing a 3D convolution block, with each token maintaining the 3D structure. To address the inadaptability of traditional Mamba for high-dimensional hyperspectral scenarios, a 3D-Spectral-Spatial Selective Scanning (3DSS) mechanism is customized as the core component of 3DMB to achieve spectral-spatial sequence modeling. The 3DSS first performs pixel-wise sequence flattening on each 3D hyperspectral token along the spectral and spatial dimensions, then introduces the S6 model to conduct selective scanning to facilitate interactions among adjacent pixels. Five scanning routes are constructed to investigate the impact of dimension prioritization, including spectral-priority, spatial-priority, cross spectral-spatial, cross spatial-spectral, and parallel spectral-spatial. Multiple 3DMBs are stacked to extract comprehensive spectral-spatial semantic features, followed by a classifier for final classification. Compared to existing methods, the proposed 3DSS-Mamba exhibits superior capabilities in capturing global spectral-spatial contextual dependencies with greater computational efficiency.
\begin{figure*}[!htb]
	\centering
	\includegraphics[width=1\textwidth]{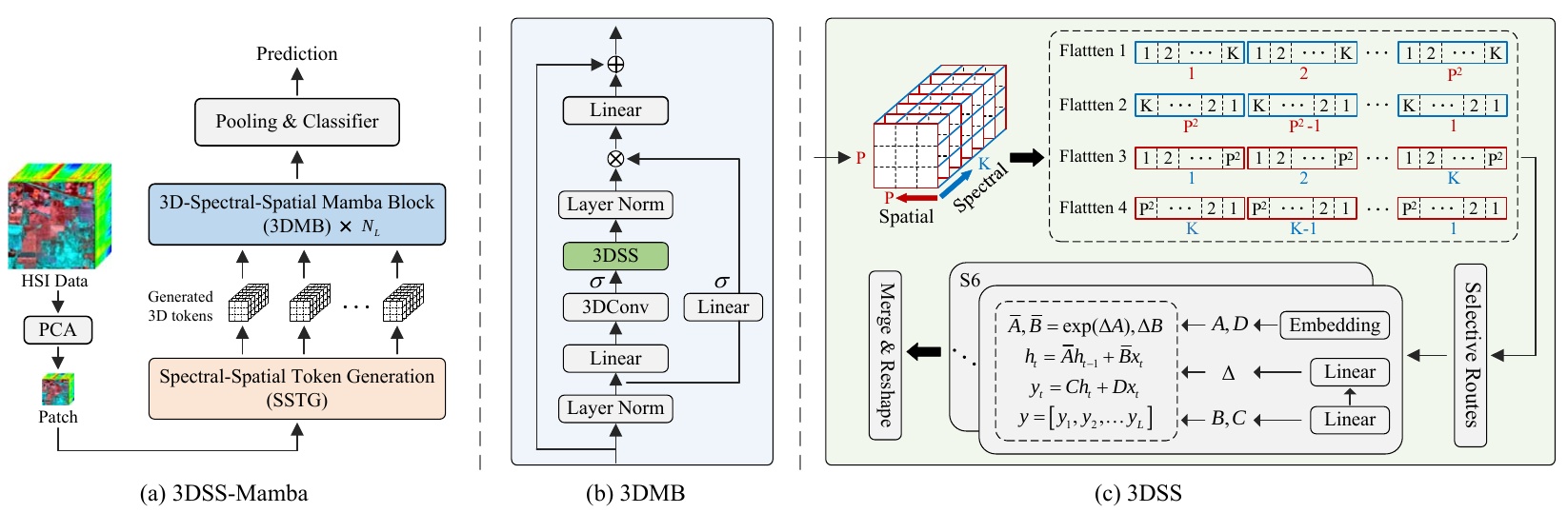}
	\caption{(a) The overall architecture of the proposed 3D-Spectral-Spatial Mamba (3DSS-Mamba) for HSI classification, which consists of a Spectral-Spatial Token Generation module (SSTG), ${N_L}$ stacked 3D-Spectral-Spatial Mamba Blocks (3DMB), and a classifier module; (b) The structural flow of proposed 3D-Spectral-Spatial Mamba Block (3DMB); (c) The computational procedure of proposed 3D-Spectral-Spatial Selective Scanning (3DSS).}
\label{fig:Frame}
\end{figure*}

The main contributions are summarized as follows.
\begin{itemize}

\item A novel 3D-Spectral-Spatial Mamba (3DSS-Mamba) framework based on the State Space Model is proposed for HSI classification, which can explicitly establish long-range spectral-spatial contextual dependencies with linear computational complexity.

\item A 3D-Spectral-Spatial Selective Scanning (3DSS) mechanism tailored for high-dimensional hyperspectral scenarios is introduced. By performing pixel-wise selective scanning on 3D hyperspectral tokens along the spectral and spatial dimensions, the spectral reflectance and spatial regularity can be adequately explored from the sequence modeling perspective.

\item Extensive experiments are verified on three public hyperspectral datasets. The results indicate the effectiveness and superiority of the proposed method.

\end{itemize}

The remaining sections of this paper are organized as follows. Sec.~\ref{sec:dnn} probides a comprehensive description of the proposed method. Sec.~\ref{sec:simu} outlines the experimental results and analyses. Conclusions and future work are discussed in Sec.~\ref{sec:conc}

\section{The proposed network}
\label{sec:dnn}
This section commences with the preliminaries associated with State Space Models (SSMs). Following this, we investigate a novel 3D-Spectral-Spatial Selective Scanning (3DSS) mechanism specifically tailored for the three-dimensional HSI data, followed by the establishment of 3D-Spectral-Spatial Mamba block (3DMB). Building upon these sub-modules, the overall architecture of the proposed 3DSS-Mamba framework for HSI classification is meticulously introduced.

\subsection{Preliminaries}
\noindent{\textbf{State Space Models (SSMs).}}
The concept of SSMs originates from continuous linear time-invariant systems. Taking a one-dimensional signal $x(t) \in \mathbb{R}$ as input, SSMs are dedicated to mapping it into a sequence $y(t) \in \mathbb{R}$ via an intermediate hidden state $h(t) \in \mathbb{R}^{N}$. Formally, this procedure can be formulated through the following linear ordinary differential equation (ODE),
\begin{equation}\label{}
  \begin{aligned}
    h'\left( t \right) &= {\bf{A}}h\left( t \right) + {\bf{B}}x\left( t \right), \\
    y\left( t \right) &= {\bf{C}}h\left( t \right),
  \end{aligned}
\label{eq:hy}
\end{equation}
where $\mathbf{A} \in \mathbb{R}^{N \times N}$ denotes the state transition matrix, and $\mathbf{B} \in \mathbb{R}^{N \times 1}, \mathbf{C} \in \mathbb{R}^{N \times 1}$ represent the projection matrices.

The continuous-time system delineated by Eq.~(\ref{eq:hy}) generally encounters challenges when integrating into discrete sequence-based deep models. To this end, the zero-order hold (ZOH) technique with a time-scale parameter $\Delta$ is subsequently employed to facilitate a straightforward discretization step, which converts the continuous parameters $\mathbf{A}$ and $\mathbf{B}$ into their discrete counterparts $\overline{\mathbf{A}}$ and $\overline{\mathbf{B}}$,
\begin{equation}\label{}
  \begin{aligned}
    \overline{\mathbf{A}} & =\exp (\Delta \mathbf{A}), \\
    \overline{\mathbf{B}} & =(\Delta \mathbf{A})^{-1}(\exp (\Delta \mathbf{A})-\mathbf{I}) \cdot \Delta \mathbf{B} \\
                          & \approx(\Delta \mathbf{A})^{-1}(\Delta \mathbf{A})(\Delta \mathbf{B}) \\
                          & =\Delta \mathbf{B},
  \end{aligned}
\label{eq:AB}
\end{equation}

After discretization, the discretized SSM system can be formulated as follows
\begin{equation}\label{}
  \begin{aligned}
    h_{t} & =\overline{\mathbf{A}} h_{t-1}+\overline{\mathbf{B}} x_{t}, \\
    y_{t} & =\mathbf{C} h_{t},
  \end{aligned}
\label{eq:SSM}
\end{equation}

To enhance computational efficiency and scalability, the convolution operation $*$ is harnessed to expedite the linear recurrence process outlined above. Consequently, the ultimate output can be synthesized as
\begin{equation}\label{}
  \begin{aligned}
    \overline{\mathbf{K}} & =\left(\mathbf{C} \overline{\mathbf{B}}, \mathbf{C} \overline{\mathbf{A B}}, \ldots, \mathbf{C \overline{A}}^{\mathbf{L}-1} \overline{\mathbf{B}}\right), \\
    \mathbf{y} & =\mathbf{x} * \overline{\mathbf{K}},
  \end{aligned}
\end{equation}
where $L$ denotes the length of input sequence, and $\overline{\mathbf{K}} \in \mathbb{R}^{L}$ serves as the structured convolutional kernel.

\noindent{\textbf{Selective State Space Models (S6).}} Traditional SSMs predominantly rely on the simplifying assumption of linear time-invariant, which enjoy the advantage of linear time complexity but struggle to capture the contextual information within input sequences. To break this limitation, Mamba \cite{gu2023mamba} implements the Selective State Space Models (S6) to achieve the interactions between sequential states. Different from traditional SSMs integrated with static parameterization, S6 allows the projection matrices to be modified as input-dependent, which achieves selective attention on each sequence unit. Concretely, the parameters $\mathbf{B} \in \mathbb{R}^{B \times L \times N}$, $\mathbf{C} \in \mathbb{R}^{B \times L \times N}$ and $\boldsymbol{\Delta} \in \mathbb{R}^{B \times L \times D}$ are dynamically calculated from the input sequence $x \in \mathbb{R}^{B \times L \times D}$.

\begin{figure*}[!htb]
	\centering
	\includegraphics[width=1\textwidth]{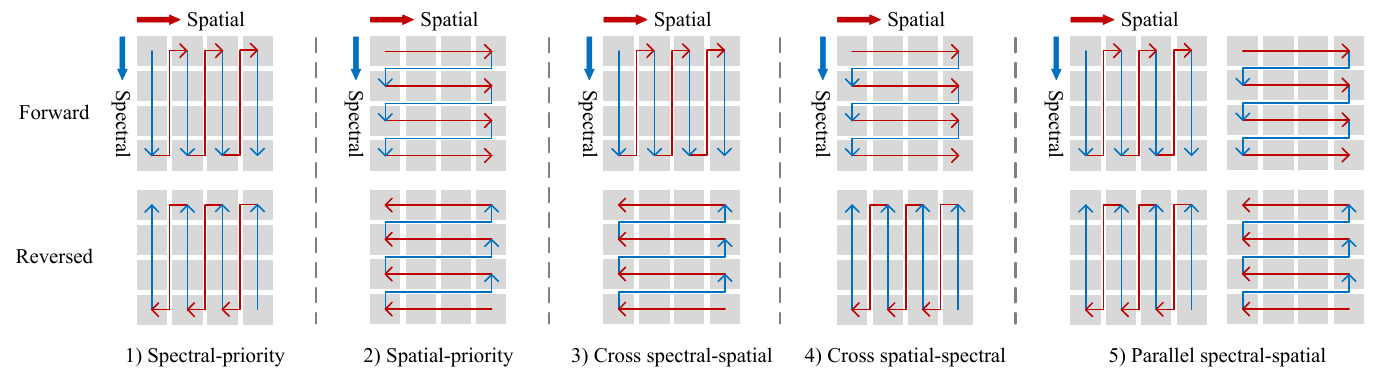}
	\caption{Five flattening routes are constructed to explore the impact of dimension prioritization.}
\label{fig:Routes}
\end{figure*}
\subsection{3D-Spectral-Spatial Selective Scanning}
\label{sec:3DSS}
The original Mamba processes data along a specific orientation, which is effectively employed for the causal modeling of 1-D input sequences. To accommodate vision tasks involving 2D-spatial awareness, recent VMamba \cite{liu2024vmamba} introduces a 2D-selective-scanning technique. The cross-scanning mechanism rearranges tokens along spatial dimensions and then transmits them into the S6 model for sequence modeling. Although the above scanning techniques have demonstrated commendable application in language data and natural image, they may encounter substantial challenges when adapting to 3D hyperspectral data that exhibit inherent visual spatial and continuous spectral characteristics. To address this issue, this paper proposes a 3D-Spectral-Spatial Selective Scanning (3DSS) module, which performs pixel-wise sequential scanning for 3-D hyperspectral input to achieve global spectral-spatial relationship modeling.

Inspired by VMamba \cite{liu2024vmamba}, the developed 3DSS is depicted in Fig.~\ref{fig:Frame}(c), primarily comprising two stages: 3D-spectral-spatial sequence flattening and selective scanning with S6 mechanism.

\noindent{\textbf{3D-Spectral-Spatial Sequence Flattening.}}
Unlike conventional 2D scanning that solely emphasizes spatial information, 3DSS performs pixel-wise sequential scanning for 3D hyperspectral tokens along both spectral and spatial dimensions, generating the flattened forward 1-D sequence. To adequately capture spectral-spatial contextual details, 3DSS additionally performs flipping operations on the forward sequence to enable bidirectional sequence scanning.

To explore the impact of dimension prioritization, five flattening routes are delineated as illustrated in Fig.~\ref{fig:Routes}: 1) Spectral-priority: initially unfolds the 3D hyperspectral token along the spectral dimension and then arranges them in spatial order. Given the input of hyperspectral tokens $F=\left\{S_{1}, S_{2}, \ldots, S_{M}\right\}, S_{i} \in \mathbb{R}^{P \times P \times K}$, where $P$ and $K$ denote the patch size and spectral band number of token cube, respectively, the flattening process for each 3D token cube $S_{i}$ can be formulated as
\begin{equation}\label{}
\begin{aligned}
    {\mathcal{S}_{i}}^{\text {spe-fwd}} & =\left[\left[\mathcal{S}_{i, 1}^{1}, \ldots, \mathcal{S}_{i, 1}^{K}\right], \ldots,\left[\mathcal{S}_{i, P^{2}}^{1}, \ldots, \mathcal{S}_{i, P^{2}}^{K}\right]\right] \\
    {\mathcal{S}_{i}}^{\text {spe-rvs}} & =\Phi_{revert}\left({\mathcal{S}_{i}}^{\text {spe-fwd}}\right),
\end{aligned}
\end{equation}
where ${\mathcal{S}_{i}}^{\text {spe-fwd}} \in \mathbb{R}^{1 \times\left(P^{2} \cdot K\right)}$ indicates the generated forward sequence, and ${\mathcal{S}_{i}}^{\text {spe-rvs}} \in \mathbb{R}^{1 \times\left(P^{2} \cdot K\right)}$ represents the reversed sequence with flipping function ${{\Phi _{revert}}}$. As a result, the bidirectional sequences driven by Spectral-priority can be expressed as ${{\cal S}_i}^{seq} = \left\{ {{{\cal S}_i}^{\text{spe-fwd}},{{\cal S}_i}^{\text{spe-rvs}}} \right\}$;
2) Spatial-priority: first organizes the token cube by spatial location and then stacks them band by band. The reversed sequence is constructed with flipping operations. Technically, the flattened sequences can be generated by
\begin{equation}\label{}
  \begin{aligned}
    {\mathcal{S}_{i}}^{\text {spa-fwd}} & =\left[\left[S_{i, 1}^{1}, \ldots, \mathcal{S}_{i, P^{2}}^{1}\right], \ldots,\left[\mathcal{S}_{i, 1}^{K}, \ldots, \mathcal{S}_{i, P^{2}}^{K}\right]\right] \\
    {\mathcal{S}_{i}}^{\text {spa-rvs}} &=\Phi_{revert}\left({\mathcal{S}_{i}}^{\text {spa-fwd}}\right).
\end{aligned}
\end{equation}
Similarly, the ultimate bidirectional sequences guided by Spectral-priority is ${{\cal S}_i}^{seq} = \left\{ {{{\cal S}_i}^{\text{spa-fwd}},{{\cal S}_i}^{\text{spa-rvs}}} \right\}$;
3) Cross spectral-spatial: a hybrid pattern integrating forward Spectral-priority and reversed Spatial-priority, which can be expressed as ${{\cal S}_i}^{seq} = \left\{ {{{\cal S}_i}^{\text{spe-fwd}},{{\cal S}_i}^{\text{spa-rvs}}} \right\}$;
4) Cross spatial-spectral: a hybrid pattern combining forward Spatial-priority and reversed Spectral-priority, i.e., ${{\cal S}_i}^{seq} = \left\{ {{{\cal S}_i}^{\text{spa-fwd}},{{\cal S}_i}^{\text{spe-rvs}}} \right\}$;
5) Parallel spectral-spatial: integrates both forward and reversed Spatial-priority and Spectral-priority routes. The generated sequences can be represented as ${{\cal S}_i}^{seq} = \left\{ {{{\cal S}_i}^{\text{spa-fwd}},{{\cal S}_i}^{\text{spa-rvs}}},{{{\cal S}_i}^{\text{spe-fwd}},{{\cal S}_i}^{\text{spe-rvs}}} \right\}$. These five routes facilitate pixel interactions among adjacent spatial and spectral positions in different dimension priority, and their effectiveness will be analyzed and compared in the experimental section.

After complimenting the flattening operation following the preset route, the generated sequences ${S_i}^{seq}$ are transmitted into the subsequent S6 model for sequence modeling.

\noindent{\textbf{Selective Scanning with S6 Model.}}
The selective scanning model S6 \cite{gu2023mamba} maintains the advantages of dynamic weights (i.e., selectivity) and linear computational complexity. Building on this, the S6 model is extended to multi-sequence hyperspectral scenario for learning spectral-spatial sequence modeling expression. Specifically, we devise multiple parallel S6 models to independently process input sequences, and eventually merge the resultant to form the output response.

Taking Spatial-priority route ${{\cal S}_i}^{seq} = \left\{ {{{\cal S}_i}^{\text{spa-fwd}},{{\cal S}_i}^{\text{spa-rvs}}} \right\}$ as example, the scanning procedure for individual sequence can be formulated as
\begin{equation}\label{}
\begin{aligned}
    {Y_{i}}^{\text {spa-fwd}} &= {\Phi _{S6\text{-}fwd}}\left( {{{\cal S}_i}^{{\text {spa-fwd}}}} \right)\\
    {Y_{i}}^{\text {spa-rvs}} &= {\Phi _{S6\text{-}rvs}}\left( {{{\cal S}_i}^{{\text {spa-rvs}}}} \right),
\end{aligned}
\end{equation}
where ${\Phi _{S6\text{-}*}}$ represents the S6 model, with the detailed computation referenced in Eq.~(\ref{eq:SSM}). After scanning, these generated one-dimensional mapping sequences are reshaped into 3D structure and subsequently merged.
\begin{equation}\label{}
  {Y_i} = {\Phi _{merge}}\left( {{Y_{i}}^{\text{spa-fwd}},{\Phi _{revert}}\left( {{Y}_i^{{\text{spa-rvs}}}} \right)} \right)
\end{equation}

As a result, the ultimate transformed output tokens can be expressed as $F^{\text {out }}=\left\{Y_{1}, Y_{2}, \ldots, Y_{M}\right\}, Y_{i} \in \mathbb{R}^{P \times P \times K}$.

\subsection{3D-Spectral-Spatial Mamba Block}
The 3D-Spectral-Spatial Mamba Block (3DMB) takes the 3DSS mechanism as its core computing unit, with the objective of capturing global spectral-spatial semantic information. The detailed structure is illustrated in Fig.~\ref{fig:Frame}(b).

Specifically, the 3DMB commences with a normalization layer to enhance the model training stability. Following this, two parallel linear embedding layers are stacked, with one branch followed by an activation function for gating signal generation, and the other branch undergoes a 3D convolution operation with kernel $1 \times 1 \times 1$. The procedure can be formulated as
\begin{equation}\label{}
  z = \sigma \left( {{\Phi _{linear}}\left( {{\Phi _{norm}}\left( {T} \right)} \right)} \right),
\end{equation}
\begin{equation}\label{}
  F = \sigma \left( {{\Phi _{3DConv}}\left( {{\Phi _{linear}}\left( {{\Phi _{norm}}\left( {T} \right)} \right)} \right)} \right),
\end{equation}
where $T \in \mathbb{R}^{M \times P \times P \times K}$ denotes the input tokens, $\sigma$ denotes the Silu \cite{elfwing2018sigmoid} activation operation. After this stage, the generated $F$ passes through the pivotal 3DSS mechanism, executing selective scanning as previously described. Subsequently, the output of 3DSS undergoes layer normalization and a gating operation. Finally, the features are transmitted to the ultimate linear layer, followed by a residual connection.
\begin{equation}\label{}
  R = {\Phi _{linear}}\left( {{\Phi _{norm}}\left( {3DSS\left( F \right)} \right) \otimes z} \right) + T.
\end{equation}

Notably, the 3DMB enjoys linear computation complexity benefiting from the 3DSS mechanism, allowing for more stackings with similar budgets compared to the Transformer.
\begin{figure}[tb]
    \centering
    \includegraphics[width=0.38\textwidth]{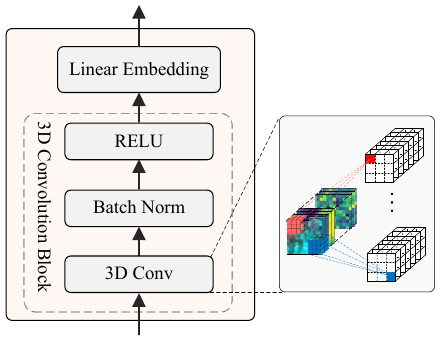}
    \caption{The detail structure of Spectral-Spatial Token Generation module (SSTG).}
    \label{fig:SSTG}
\end{figure}

\subsection{3D-Spectral-Spatial Mamba: Overview}
The architecture of the proposed 3D-Spectral-Spatial Mamba (3DSS-Mamba) for HSI classification is illustrated in Fig~\ref{fig:Frame}(a). It consists of a Spectral-Spatial Token Generation module (SSTG), multiple stacked 3D-Spectral-Spatial Mamba blocks (3DMB), and a prediction module. Initially, the cropped patch cube is fed into the SSTG to acquire a series of 3D spectral-spatial tokens. Subsequently, the generated tokens are input into the stacked 3DMB to capture discriminative spectral-spatial semantic representations. Ultimately, the extracted spectral-spatial features are transmitted to the prediction module to accomplish classification.

Assume that the original hyperspectral data is expressed as $I \in {\mathbb{R} ^{H \times W \times V}}$, where $H$ and $W$ represent the spatial dimensions, and $V$ denotes the spectral dimension. To mitigate the potential Hughes phenomenon caused by high dimensionality, dimensionality reduction is initially conducted on the original hyperspectral data through principal component analysis (PCA) \cite{renard2008denoising}. The modified image is represented as ${I_{PCA}} \in {\mathbb{R}^{H \times W \times d}}$, where $d$ refers to the reduced spectral dimension. Given that the adjacent pixels can supplement spatial information for the central pixel, the modified image is further divided into a series of 3-D patch cubes $\left\{ {{x_i} \in {\mathbb{R}^{B \times B \times d}}} \right\}_{i = 1}^{H \times W}$ as input for 3DSS-Mamba, with the labels determined by the central pixel of each patch.

The developed Spectral-Spatial Token Generation module (SSTG) is constructed by a 3D convolution block and an embedding operation, which projects the HSI patch cube into spectral-spatial tokens. The detailed structure is shown in Fig.~\ref{fig:SSTG}. Taking the cropped pixel-wise patch cube $x \in {\mathbb{R}^{B \times B \times d}}$ as input, the tokenization process can be formulated as
\begin{equation}\label{}
  T = {\Phi _{embed}}\left( {{\Phi _{3DConv}}\left( x \right)} \right).
\end{equation}
where $T \in \mathbb{R}^{M \times P \times P \times K}$ denotes the generated spectral-spatial token. The 3D convolution block consists of a 3D convolution layer, a batch normalization layer, and a ReLU activation function. The embedding operation involves a linear layer for dimension transformation.

Subsequently, the generated tokens are fed into by ${N_L}$ stacked 3D-Spectral-Spatial Mamba blocks (3DMB) for spectral-spatial semantic extraction. This procedure can be iteratively delineated as follows
\begin{equation}\label{}
  \begin{aligned}
    &{R^j} = \Phi _{3DMB}^j\left( {{T^j}} \right)\\
    &{T^{j + 1}} = {R^j},
  \end{aligned}
\end{equation}
where $\Phi _{3DMB}^j$ signifies the $j$-$th$ 3DMB block, and ${R^j}\in \mathbb{R}^{M \times P \times P \times K}$ represents the corresponding output.

After completing the 3DMB modeling, the spectral-spatial feature ${R^{{N_L}}}$ undergoes an average pooling operation, and then passes through the classifier ${\Phi _{classifier}}$ comprised of multilayer perceptron layers to yield the ultimate classification results
\begin{equation}\label{}
  pred = {\Phi _{classifier}}\left( {{\Phi _{avg}}\left( {{R^{{N_L}}}} \right)} \right).
\end{equation}

\section{Performance evaluation}
\label{sec:simu}

\subsection{Datasets Description}
To illustrate the classification capabilities of the proposed 3DSS-Mamba, three publicly available HSI databases are utilized for comprehensive evaluation, including Pavia University, Indian Pines, and Houston 2013. Detailed descriptions are provided below.

\subsubsection{Pavia University} The dataset was collected by the Reflective Optics System Imaging Spectrometer (ROSIS) over Pavia, Northern Italy. The imaging wavelength of the spectrometer ranges from 0.43 to 0.86 $\mu$m. After removing the noisy bands, the dataset consists of 103 spectral bands and 610 $\times$ 340 pixels, with a spatial resolution of 1.3 m per pixel. There are totally 42776 ground sample points, categorized into 9 types including Asphalt, Gravel, etc. Table~\ref{PU_dataset} provides the division details for training and testing sets.

\subsubsection{Indian Pines} The dataset was acquired by the Airborne/Visible Infrared Imaging Spectrometer (AVIRIS) imaging an Indian pine tree over Northwestern Indiana in 1992. The imaging wavelength of the spectrometer ranges from 0.4 to 2.5 $\mu$m. The dataset encompasses 200 spectral bands and 145 $\times$ 145 pixels after removing the water absorption channels, with a spatial resolution of 20 m per pixel. There are a total of 10249 ground object pixels, representing 16 distinct categories including Alfalfa, Corn-notill, etc. The detailed data division in the experiment is described in Table~\ref{Indian_dataset}.

\subsubsection{Houston 2013} The dataset was captured by the ITRES CASI-1500 sensor over the University of Houston campus and its surrounding areas, provided by the 2013 GRSS Data Fusion Contest. The image comprises 144 spectral bands ranging in wavelength from 0.38 to 1.05 $\mu$m, and consists of 340 $\times$ 1905 pixels with a spatial resolution of 2.5 m per pixel. There are 16373 sample pixels, covering 15 challenging land cover categories. The precise splitting of training and testing data is exhibited in Table~\ref{HU2013_dataset}.
\begin{table}
\caption{\textsc{Name and Number of Samples of Each Class on the Pavia University Datasets with 5\% labeled data.}}
\centering
\renewcommand\arraystretch{1.1}
     \centerline{
     \begin{tabular}{ p{0.9cm}<{\centering}| p{3.0cm}<{\centering} p{1.0cm}<{\centering} p{1.0cm}<{\centering}}
     \hline
     \multicolumn{4}{c}{\textbf{Pavia University dataset}}\\\hline
     No.&Class Name&Train&Test \\
     \hline\rule{0pt}{8pt}
      1&Asphalt &332 &6299 					\\
      2&Meadows &932 &17717 					\\
      3&Gravel &105 &1994						\\
      4&Trees &153 &2911 						 \\
      5&Painted metal sheets &67 &1278	\\
      6&Bare soil &251 &4778\\
      7&Bitumen &67 &1263 \\
      8&Self-blocking bricks &184 &3498\\
      9&Shadows &47 &900 \\
     \hline\rule{0pt}{8pt}
       &Total  &2138 &40638		\\
     \hline
     \end{tabular}}
     \label{PU_dataset}
 \end{table}

\begin{table}
 \caption{\textsc{Name and Number of Samples of Each Class on the Indian Pines Datasets with 10\% labeled data.}}
\renewcommand\arraystretch{1.1}
 \centerline{
     \begin{tabular}{ p{0.8cm}<{\centering}| p{3.0cm}<{\centering} p{1.0cm}<{\centering} p{1.0cm}<{\centering}}
     \hline
     \multicolumn{4}{c}{\textbf{Indian Pines dataset}}\\\hline
     No.&Class Name&Train&Test\\
     \hline\rule{0pt}{8pt}
      1 &Alfalfa 		&5 &41 		\\
      2 &Corn-notill			&143 &1285 	\\
      3 &Corn-mintill		 	&83 &747 		\\
      4 &Corn      	    &24 &213 	\\
      5 &Grass-pasture-mowed 		&48 &435		\\
      6 &Grass-trees 		&73 &657		\\
      7 &Grass-pasture 		&3 &25			\\
      8 &Hay-windrowed 			&48 &430\\
      9 &Oats 			&2 &18\\
      10 &Soybean-notill			&97 &875 		\\
      11 &Soybean-mintill 		&245 &2210 	\\
      12 &Soybean-clean 		&59 &534		\\
      13 &Wheat &20 &185 		\\
      14 &Woods 			&126 &1139 			\\
      15 &Buildings 			&39 &347 			\\
      16 &Stone &6 &84 \\
     \hline\rule{0pt}{8pt}
       &Total  				&1024 &9225 \\
     \hline
     \end{tabular}}
     \label{Indian_dataset}
 \end{table}

\begin{table}
 \caption{\textsc{Name and Number of Samples of Each Class on the Houston 2013 Datasets with 10\% labeled data.}}
\renewcommand\arraystretch{1.1}
     \centerline{
     \begin{tabular}{ p{0.8cm}<{\centering}| p{3.0cm}<{\centering} p{1.0cm}<{\centering} p{1.0cm}<{\centering}}
     \hline
     \multicolumn{4}{c}{\textbf{Houston 2013 dataset}}\\\hline
     No.&Class Name&Train&Test \\
     \hline\rule{0pt}{8pt}
      1 &Healthy grass 	&125 &1238\\
      2 &Stressed grass 	&125 &1241 \\
      3 &Synthetic grass &70 &690\\
      4 &Trees 			&124 &1231 \\
      5 &Soil 			&124 &1229\\
      6 &Water &33 &321\\
      7 &Residential &127 &1255 \\
      8 &Commercial 	&124 &1231 \\
      9 &Road 		&125 &1239\\
      10&Highway&123 &1214\\
      11&Railway&123 &1222\\
      12&Parking Lot 1&123 &1220\\
      13&Parking Lot 2&47 &464\\
      14&Tennis Court&43 &423\\
      15&Running Track&66 &653 \\
     \hline\rule{0pt}{8pt}
       &Total  &1502 &14871 \\
     \hline
     \end{tabular}}
     \label{HU2013_dataset}
 \end{table}
 
\begin{figure}[tb]
    \centering
	\includegraphics[width=8cm,height=6cm]{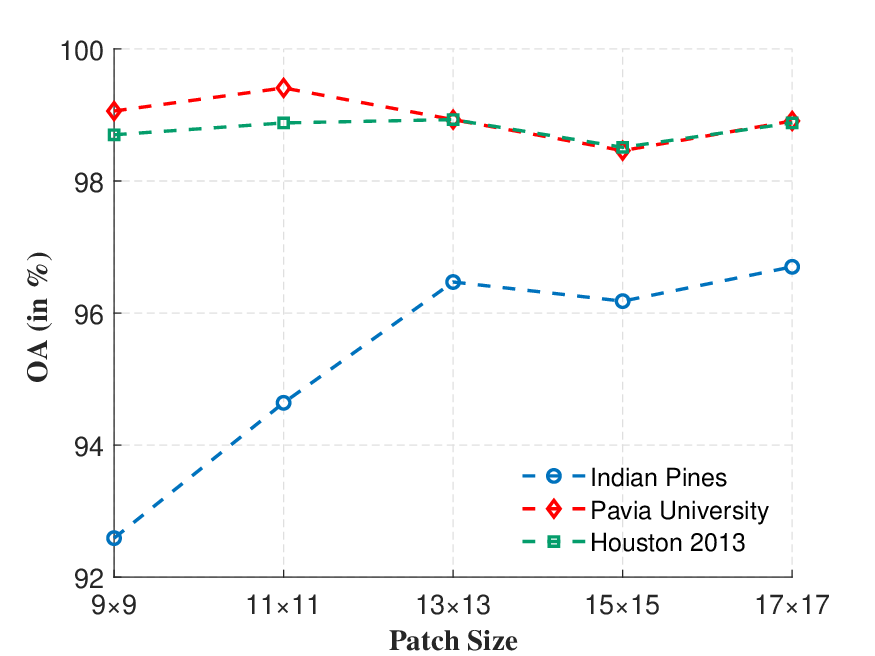}
	\caption{Sensitivity analysis for the proposed method with different sizes of input patches.}
	\label{fig:Patchsize}
\end{figure}

\begin{table*}[!htb]
\centering
\caption{\textsc{Ablation study on the accuracy metrics for different scanning routes in 3DSS.}}
\setlength\tabcolsep{8pt}
\renewcommand\arraystretch{1.1}
\begin{tabular}{lc|ccc|ccc|ccc}
\hline
\multicolumn{2}{c|}{\multirow{2}{*}{\textbf{Routes}}} & \multicolumn{3}{c|}{\textbf{Pavia University}}   & \multicolumn{3}{c|}{\textbf{Indian Pines}}       & \multicolumn{3}{c}{\textbf{Houston 2013}}        \\ \cline{3-11}
\multicolumn{2}{c|}{}                                 & OA(\%)         & AA(\%)         & Kappa          & OA(\%)         & AA(\%)         & Kappa          & OA(\%)         & AA(\%)         & Kappa          \\ \hline
\multicolumn{1}{l|}{1}   & Spectral-priority          & 98.34          & 97.12          & 97.81          & 93.21          & 84.73          & 92.25          & 97.95          & 98.07          & 97.78          \\
\multicolumn{1}{l|}{2}   & Spatial-priority           & 99.18          & 98.14          & 98.91          & 95.49          & 88.06          & 94.85          & 98.31          & 98.39          & 98.17          \\
\multicolumn{1}{l|}{3}   & Cross spectral-spatial     & 99.07          & 98.5           & 98.77          & 96.16          & 89.41          & 95.61          & 98.5           & 98.57          & 98.38          \\
\multicolumn{1}{l|}{4}   & Cross spatial-spectral     & 99.32          & 98.74          & \textbf{99.14} & 96.16          & 90.22          & 95.62          & 98.84          & \textbf{99.04} & \textbf{98.86} \\
\multicolumn{1}{l|}{5}   & Parallel spectral-spatial  & \textbf{99.34} & \textbf{98.95} & 99.12          & \textbf{96.47} & \textbf{93.09} & \textbf{95.97} & \textbf{98.93} & 98.92          & 98.85          \\ \hline
\end{tabular}
\label{tab:Routes}
\end{table*}

\begin{figure*}[htbp]
	\centering
    \begin{subfigure}{0.32\linewidth}
		\centering
		\includegraphics[width=6cm,height=4cm]{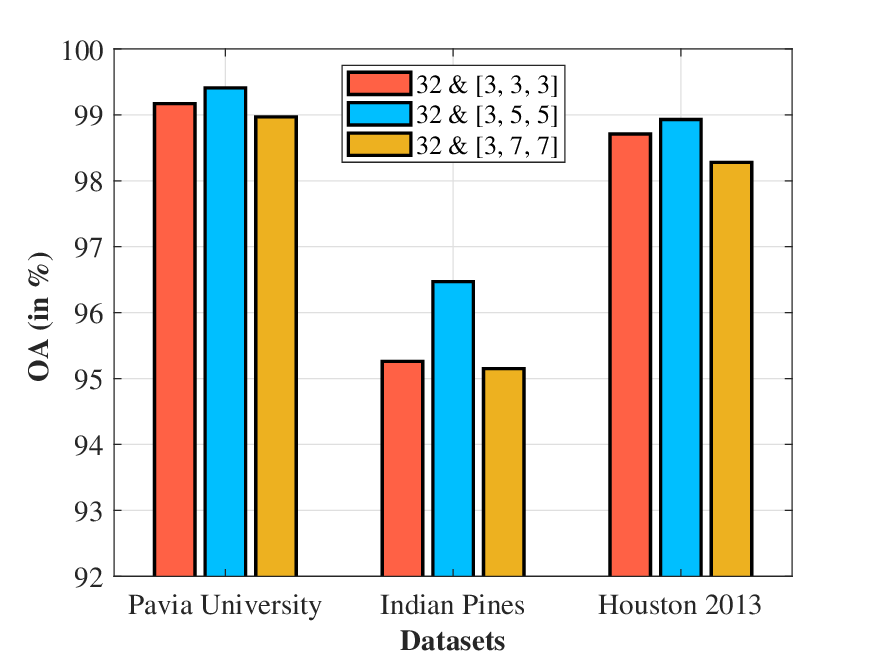}
		\caption{}
        \setlength{\belowdisplayskip}{43pt}
		\label{pseudo-I}
	\end{subfigure}
	\centering
	\begin{subfigure}{0.32\linewidth}
		\centering
		\includegraphics[width=6cm,height=4cm]{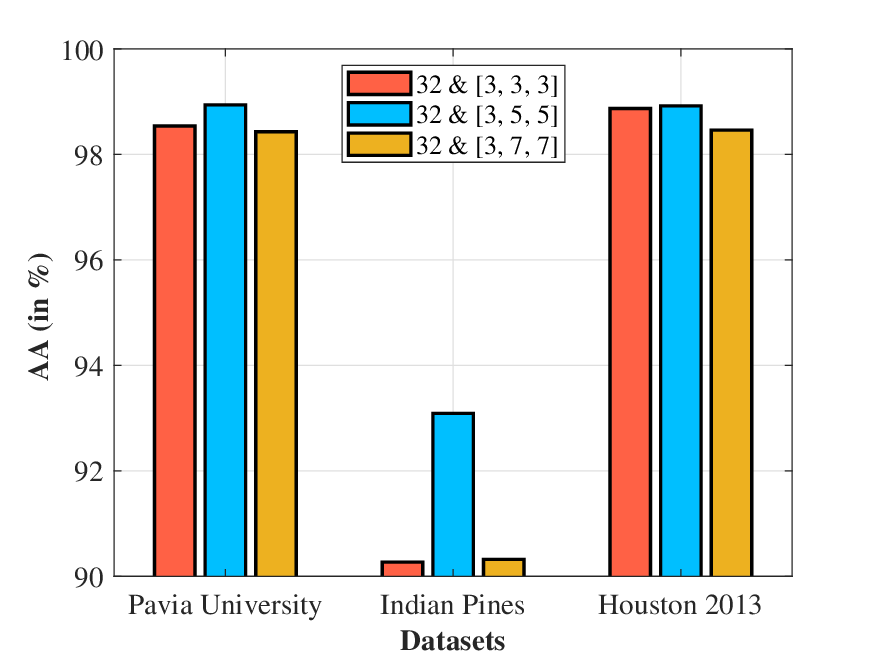}
		\caption{}
		\label{pseudo-I}
	\end{subfigure}
    \begin{subfigure}{0.3\linewidth}
		\centering
		\includegraphics[width=6cm,height=4cm]{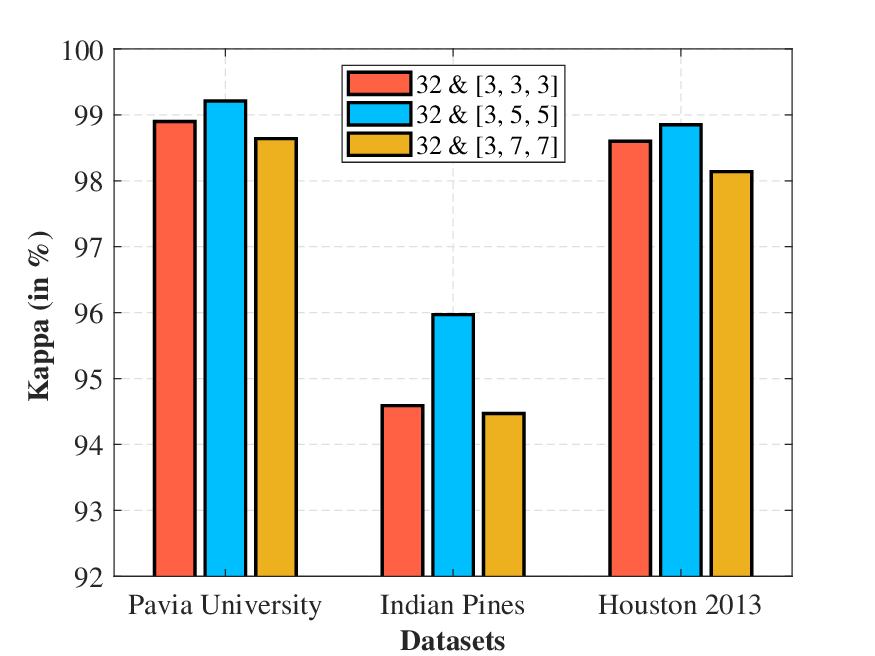}
		\caption{}
		\label{pseudo-I}
	\end{subfigure}
    \caption{Sensitivity analysis for the proposed method with different scales of 3D\_Conv kernel in SSTG in terms of OA, AA, $\kappa$.}
	\label{fig:3DConv}
\end{figure*}

\begin{figure*}[htbp]
	\centering
    \begin{subfigure}{0.32\linewidth}
		\centering
		\includegraphics[width=6cm,height=4cm]{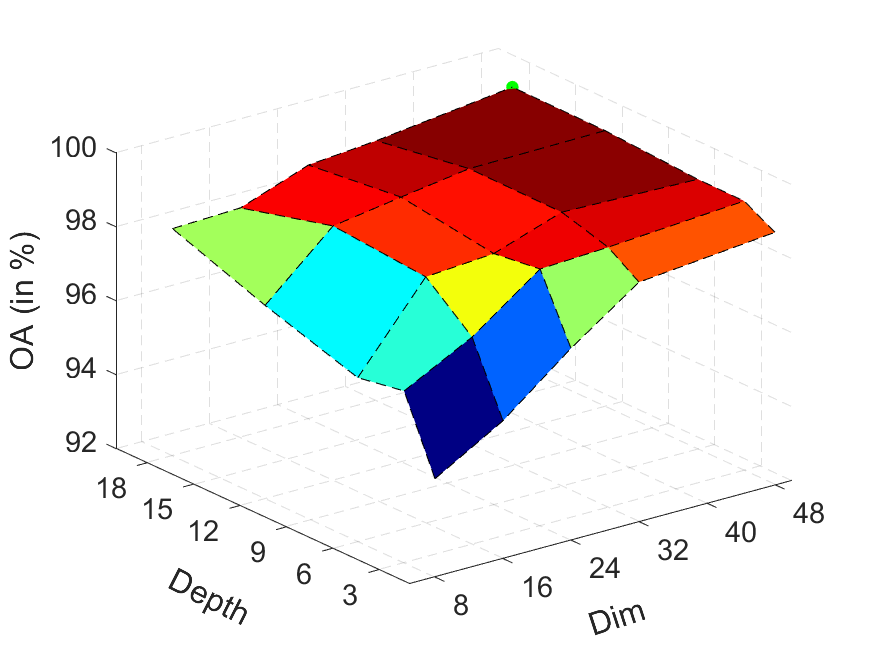}
		\caption{}
        \setlength{\belowdisplayskip}{43pt}
		\label{pseudo-I}
	\end{subfigure}
	\centering
	\begin{subfigure}{0.32\linewidth}
		\centering
		\includegraphics[width=6cm,height=4cm]{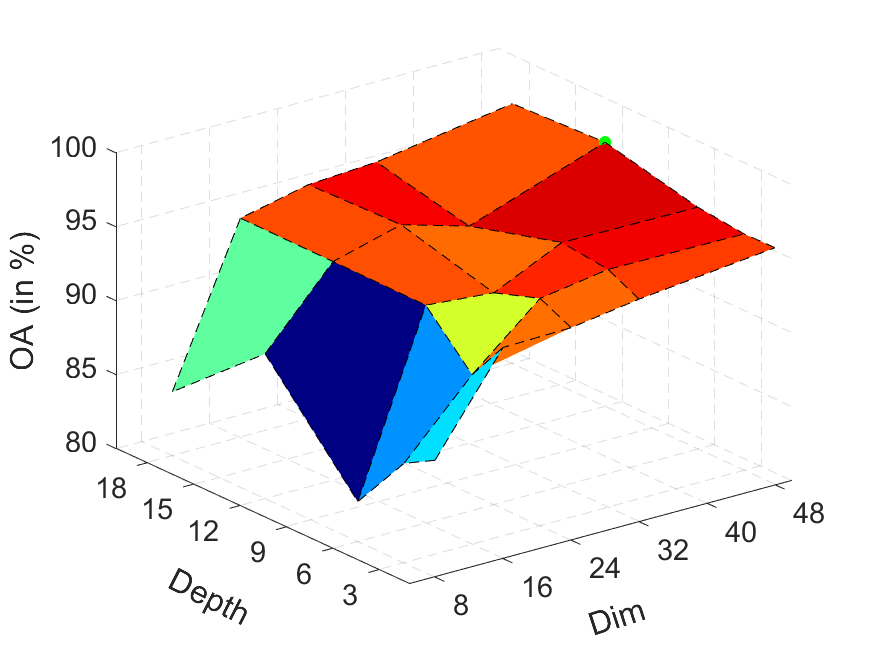}
		\caption{}
		\label{pseudo-I}
	\end{subfigure}
    \begin{subfigure}{0.3\linewidth}
		\centering
		\includegraphics[width=6cm,height=4cm]{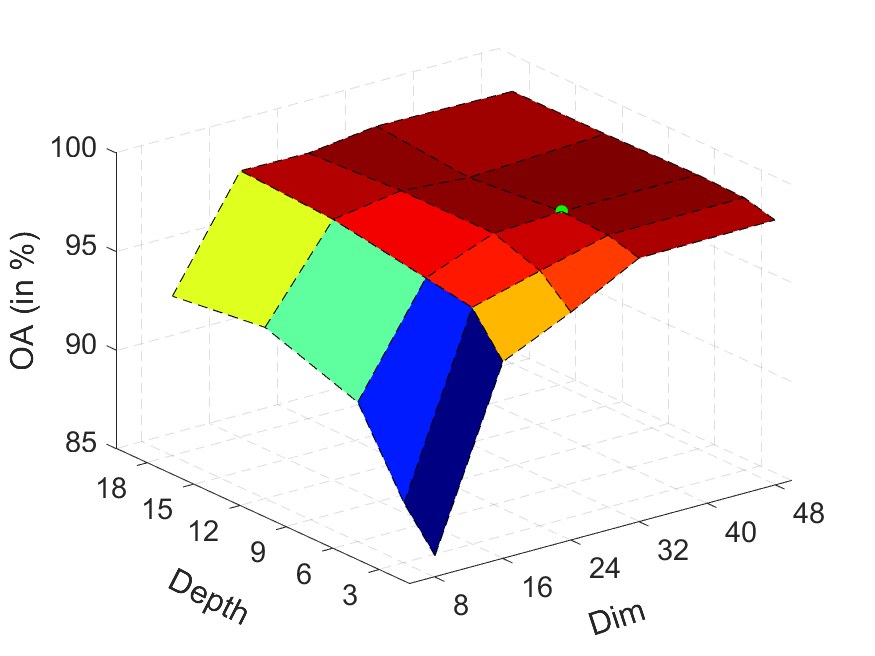}
		\caption{}
		\label{pseudo-I}
	\end{subfigure}
    \caption{Sensitivity analysis for the proposed method with different embedding dims and depths for 3DMB. (a) Pavia University. (b) Indian Pines. (c) Houston 2013.}
	\label{fig:3DMB}
\end{figure*}

\subsection{Experimental Settings}
\subsubsection{Evaluation Metrics}
Following the state-of-the-art HSI classification approaches, overall classification accuracy (OA), average classification accuracy (AA), and kappa coefficient (Kappa) are employed as the evaluation metrics. To guarantee fairness in comparison, all experiments are performed under identical experimental conditions, and the reported results are averaged over five consecutive experiments.

\subsubsection{Implementation Details}
All the experiments are implemented on the PyTorch platform with one RTX 3090Ti GPU. The training epochs and batch size are set as 100 and 64, respectively. The Adam gradient descent optimizer with learning rate 0.001 is exploited for parameter optimization. The PCA dimension for reduction is set to 30. Following the default hyperparameters in VMamba \cite{liu2024vmamba}, the state dimension and expansion ratio in the 3DSS mechanism are fixed at 16 and 2, respectively.

\subsection{Ablation Study}
\subsubsection{Effectiveness of Different Scanning Routes in 3DSS}
Acknowledging the impact of scanning dimension priority on modeling capability, this section explores the effectiveness of constructed five scanning routes, including Spectral-priority, Spatial-priority, Cross spectral-spatial, Cross spatial-spectral, and Parallel spectral-spatial. Table~\ref{tab:Routes} illustrates the classification results in terms of accuracy metrics. As can be observed, all scanning routes achieve significant classification performance, demonstrating the superiority of 3DSS in modeling global spectral-spatial contextual relationships. Comparatively, the Spatial-priority scanning route demonstrates more competitive advantages than the spectrum-prioritized mechanism. The integration of spatial and spectral information further contributes to the enhancement of classification capability. Notably, the Parallel spectral-spatial route showcases the optimal performance across all three datasets, benefiting from both the spatial and spectral priorities with bidirectional modeling. Taking the Pavia University dataset as an example, the Parallel spectral-spatial route surpasses the basic Spectral-priority by margins of 1.0\%, 1.83\%, and 1.31\% for OA, AA, and Kappa, respectively. As a result, the Parallel spectral-spatial route is selected for subsequent experiments.
\begin{table*}[!htb]
\centering
\caption{\textsc{Classification Accuracies of the Compared Methods in Terms of OA, AA, $\kappa$, and the Accuracies of Each CLasses for the Pavia University Dataset. The Best Accuracies are Presented in Bold.}}
\setlength\tabcolsep{6.5pt}
\renewcommand\arraystretch{1.1}
\begin{tabular}{c|c|ccc|cccc|c}
\hline
\multirow{2}{*}{\textbf{Class}} & \multicolumn{1}{l|}{\textbf{Conventional}} & \multicolumn{3}{c|}{\textbf{CNN-based Methods}} & \multicolumn{4}{c|}{\textbf{Transformer-based Methods}}       & \multirow{2}{*}{\textbf{3DSS-Mamba}} \\ \cline{2-9}
                                & SVM                                        & 1D-CNN      & 2D-CNN      & 3D-CNN              & VIT(Pixel) & VIT(Patch)     & SF             & HSI-BERT       &                                      \\ \hline
1                               & 91.63                                      & 83.11       & 96.31       & 95.94               & 82.87      & 92.46          & 94.75          & 97.63          & \textbf{99.33}                       \\
2                               & 97.56                                      & 92.78       & 99.37       & 99.71               & 94.00      & 96.53          & 98.14          & \textbf{99.93} & 99.34                                \\
3                               & 73.90                                      & 60.03       & 82.22       & 89.76               & 75.91      & 93.73          & 84.70          & 86.51          & \textbf{95.09}                       \\
4                               & 92.10                                      & 85.57       & 94.35       & 97.36               & 83.39      & 95.64          & 96.81          & \textbf{97.53} & 95.91                                \\
5                               & 98.51                                      & 97.77       & 99.85       & \textbf{100.0}      & 99.14      & \textbf{100.0} & 99.22          & 99.61          & 99.69                                \\
6                               & 86.04                                      & 70.77       & 93.65       & 97.41               & 51.49      & 98.47          & 94.37          & 93.05          & \textbf{99.43}                       \\
7                               & 84.09                                      & 44.21       & 89.17       & \textbf{96.69}      & 41.34      & 94.62          & 84.56          & 95.80          & 95.88                                \\
8                               & 90.25                                      & 66.38       & 88.73       & 93.92               & 76.48      & \textbf{98.60} & 92.65          & 97.86          & 96.20                                \\
9                               & 99.22                                      & 80.89       & 97.57       & \textbf{100.0}      & 99.34      & 98.11          & \textbf{100.0} & 99.67          & 97.22                                \\ \hline
OA (\%)                         & 92.75                                      & 82.68       & 95.77       & 97.62               & 82.77      & 96.18          & 96.60          & 97.61          & \textbf{98.48}                       \\
AA (\%)                         & 90.37                                      & 75.72       & 93.47       & 96.75               & 78.22      & 96.46          & 93.91          & 96.40          & \textbf{97.56}                       \\
Kappa                           & 90.35                                      & 76.92       & 94.37       & 96.84               & 76.77      & 94.97          & 94.17          & 96.82          & \textbf{97.98}                       \\ \hline
\end{tabular}
\label{tab:PU}
\end{table*}

\begin{figure*}[!htb]
	\centering
	\includegraphics[width=0.8\textwidth]{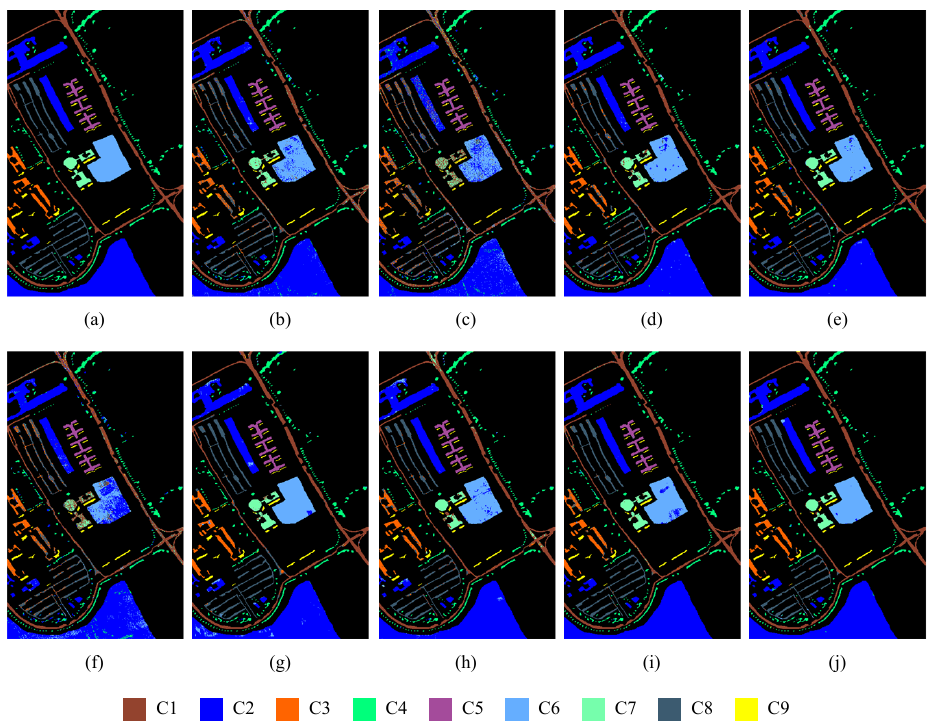}
	\caption{Classification maps obtained by the compared methods on the Pavia University dataset. (a) Reference map. (b) SVM. (c) 1D-CNN. (d) 2D-CNN. (e) 3D-CNN. (f) VIT(Pixel). (g) VIT(Patch). (h) SF. (i) HSI-BERT. (j) The proposed 3DSS-Mamba.}
\label{fig:PU}
\end{figure*}

\subsection{Parameter Analysis}
In this section, a series of experiments are carried out to analyze and determine the optimal parameters for 3DSS-Mamba, including the input patch sizes, the 3D convolution scales in SSTG, and the embedding dims and depths for 3DMB. 

\subsubsection{Different Input Patch Sizes}
Fig.~\ref{fig:Patchsize} depicts the classification performance across different input patch sizes, ranging from $9 \times 9$ to $17 \times 17$ with a growth interval of 2. As observed from the figures, the Indian Pines and Houston 2013 exhibit a similar variation tendency of consistently increasing and then decreasing, with the maximum peak at the identified point $13 \times 13$. For the Pavia University dataset, accuracy diminishes as the input patch size increases to 11. Accordingly, a patch size of $13 \times 13$ is leveraged for the Indian Pines and Houston 2013 datasets, and $11 \times 11$ is employed for the Pavia University dataset.
\begin{table*}[!htb]
\centering
\caption{\textsc{Classification Accuracies of the Compared Methods in Terms of OA, AA, $\kappa$, and the Accuracies of Each CLasses for the Indian Pines Dataset. The Best Accuracies are Presented in Bold.}}
\setlength\tabcolsep{6.5pt}
\renewcommand\arraystretch{1.1}
\begin{tabular}{c|c|ccc|cccc|c}
\hline
\multirow{2}{*}{\textbf{Class}} & \textbf{Conventional} & \multicolumn{3}{c|}{\textbf{CNN-based Methods}} & \multicolumn{4}{c|}{\textbf{Transformer-based Methods}} & \multirow{2}{*}{\textbf{3DSS-Mamba}} \\ \cline{2-9}
                                & SVM                   & 1D-CNN    & 2D-CNN           & 3D-CNN           & VIT(Pixel) & VIT(Patch)     & SF             & HSI-BERT &                                      \\ \hline
1                               & 20.73                 & 15.22     & 76.09            & 95.65            & 0.00       & 70.73          & \textbf{100.0} & 48.78    & 78.05                                \\
2                               & 73.07                 & 64.43     & \textbf{92.16}   & 88.31            & 36.65      & 59.22          & 78.37          & 78.44    & 90.27                                \\
3                               & 62.99                 & 50.48     & 87.23            & 84.94            & 0.94       & 60.51          & 89.29          & 80.58    & \textbf{93.04}                       \\
4                               & 50.70                 & 28.69     & 64.98            & 87.76            & 4.23       & 90.14          & 81.22          & 51.17    & \textbf{93.43}                       \\
5                               & 92.64                 & 76.60     & 96.48            & 96.48            & 20.46      & 64.14          & 89.89          & 91.24    & \textbf{97.70}                       \\
6                               & 94.90                 & 85.89     & \textbf{98.77}   & 96.16            & 94.22      & 97.72          & 97.41          & 96.19    & 97.26                                \\
7                               & 76.00                 & 32.14     & 32.14            & 78.57            & 0.00       & \textbf{100.0} & 84.00          & 20.00    & 96.00                                \\
8                               & 96.63                 & 86.40     & \textbf{100.0}   & \textbf{100.0}   & 99.53      & 97.91          & \textbf{100.0} & 98.83    & 99.30                                \\
9                               & 33.33                 & 15.00     & 30.00            & \textbf{70.00}   & 0.00       & 0.00           & 27.78          & 0.00     & 38.89                                \\
10                              & 68.69                 & 60.08     & 91.87            & 92.80            & 33.26      & 85.71          & 94.97          & 78.37    & \textbf{96.80}                       \\
11                              & 85.16                 & 69.61     & 95.93            & 91.20            & 88.96      & 92.81          & 93.44          & 92.85    & \textbf{98.91}                       \\
12                              & 64.89                 & 56.32     & \textbf{93.42}   & 88.03            & 4.12       & 90.07          & 81.65          & 62.47    & 91.39                                \\
13                              & 97.03                 & 89.27     & 97.56            & 94.15            & 82.70      & \textbf{98.38} & 87.03          & 96.19    & 95.68                                \\
14                              & 96.49                 & 79.45     & 98.81            & 99.05            & 98.24      & 97.01          & 94.82          & 94.46    & \textbf{99.65}                       \\
15                              & 55.33                 & 49.48     & 81.35            & 89.38            & 11.53      & \textbf{99.71} & 96.83          & 87.89    & 94.52                                \\
16                              & 93.93                 & 40.86     & 97.85            & \textbf{100.0}   & 91.67      & \textbf{100.0} & 95.24          & 65.06    & 84.52                                \\ \hline
OA (\%)                         & 79.82                 & 67.13     & 93.34            & 92.17            & 57.35      & 84.55          & 90.67          & 85.45    & \textbf{95.82}                       \\
AA (\%)                         & 72.03                 & 56.25     & 83.41            & 90.78            & 41.65      & 81.50          & 87.00          & 71.41    & \textbf{90.83}                       \\
Kappa                           & 76.84                 & 62.44     & 92.39            & 91.08            & 48.99      & 82.32          & 89.36          & 83.36    & \textbf{95.23}                       \\ \hline
\end{tabular}
\label{tab:IP}
\end{table*}

\begin{figure*}[!htb]
	\centering
	\includegraphics[width=0.8\textwidth]{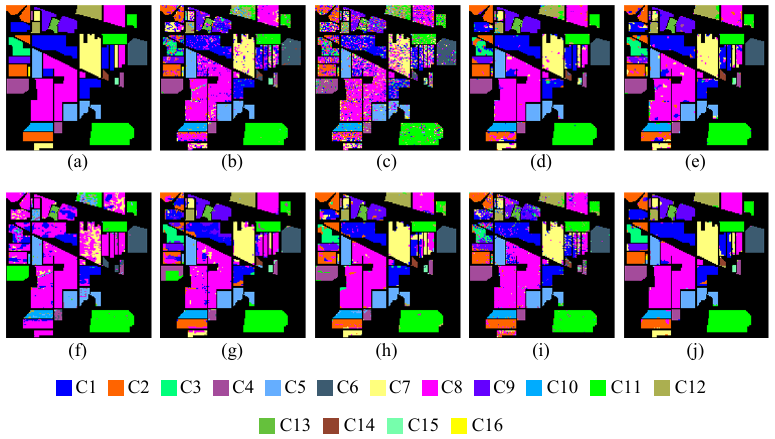}
	\caption{Classification maps obtained by the compared methods on the Indian Pines dataset. (a) Reference map. (b) SVM. (c) 1D-CNN. (d) 2D-CNN. (e) 3D-CNN. (f) VIT(Pixel). (g) VIT(Patch). (h) SF. (i) HSI-BERT. (j) The proposed 3DSS-Mamba.}
\label{fig:IP}
\end{figure*}

\subsubsection{Different Scales of 3D Convolution Kernel in SSTG}
The generated 3D spectral-spatial tokens are determined by the scale of 3D convolution kernels within the spectral-spatial token generation module. Fig.~\ref{fig:3DConv} illustrates the classification sensitivity achieved with distinct 3D kernels on the three datasets. It can be observed that appropriately increasing the scale of 3D\_Conv kernels contributes to capturing richer spectral-spatial contextual information. Based on the results, the optimal kernel scale for all three datasets is established as 32 \& {[}3, 5, 5{]}.

\subsubsection{Different Embedding Dims and Depths for 3DMB}
As the core of 3DSS-Mamba, the 3DMB module based on 3DSS scanning mechanism is iteratively stacked to achieve the extraction of global spectral-spatial semantic representations. To explore the optimal structure of 3DSS-Mamba for classification, mixed experiments are carried out by simultaneously adjusting the embedding dimensions in 3DSS and the stacked depth of 3DMB. Fig.~\ref{fig:3DMB} demonstrates the classification sensitivity on three datasets. The range of embedding dim is set to $\left\{ {8,16,24,32,48} \right\}$, and the depth covers an interval
of $\left\{ {1,3,6,12,18} \right\}$. The optimal combination is highlighted with a green point. As observed across all three datasets, lower embedding dimensions can lead to performance degradation due to underfitting. Conversely, excessively high dimensions and deeper depths provide limited accuracy improvements but computational burdens. By trading off these metrics, the proposed 3DSS-Mamba is constructed as a lightweight structure, where the embedding dimension is determined as 32, and the stacked depth is selected as 1.

\subsection{Experimental Comparison With Competitive Approaches}
\begin{figure*}[!htb]
	\centering
	\includegraphics[width=0.85\textwidth]{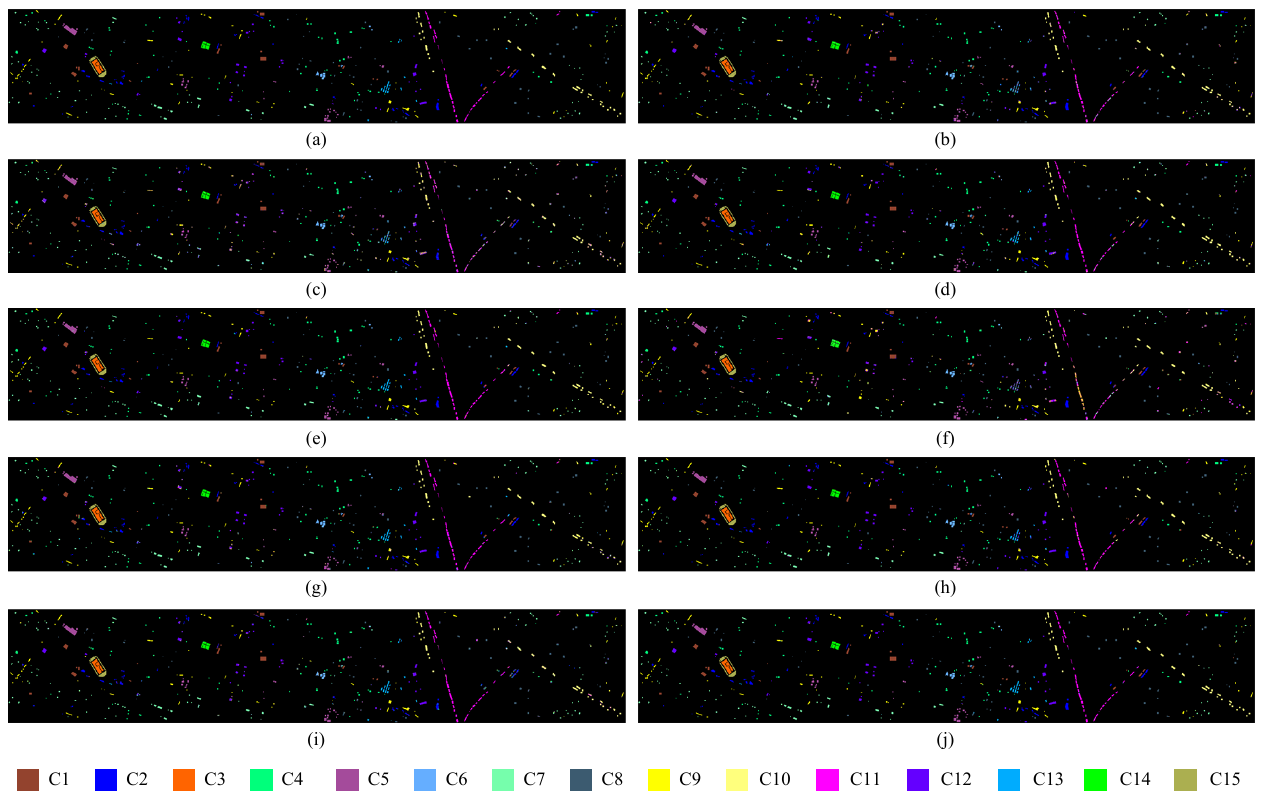}
	\caption{Classification maps obtained by the compared methods on the Houston 2013 dataset. (a) Reference map. (b) SVM. (c) 1D-CNN. (d) 2D-CNN. (e) 3D-CNN. (f) VIT(Pixel). (g) VIT(Patch). (h) SF. (i) HSI-BERT. (j) The proposed 3DSS-Mamba.}
\label{fig:HU13}
\end{figure*}

To demonstrate the effectiveness of the proposed 3DSS-Mamba, three kinds of representative HSI classification architectures are selected for comprehensive comparison, including conventional methods (SVM \cite{melgani2004classification}), CNN-based methods (1D-CNN \cite{hu2015deep}, 2D-CNN \cite{chen2016deep}, 3D-CNN \cite{zhong2017spectral}), and Transformer-based methods (VIT \cite{dosovitskiy2020image}, SF \cite{hong2022spectralformer}, HSI-BERT \cite{he2019hsi}). The quantitative accuracies (OA (\%), AA (\%), and Kappa (\%)) on the Pavia University, Indian Pines, and Houston 2013 datasets are summarized in Table~\ref{tab:PU}-\ref{tab:HU13}, with the best results highlighted in bold. Corresponding visualization maps are provided in Fig~\ref{fig:PU}-\ref{fig:HU13}.
\begin{table*}[!htb]
\centering
\caption{\textsc{Classification Accuracies of the Compared Methods in Terms of OA, AA, $\kappa$, and the Accuracies of Each CLasses for the Houston 2013 Dataset. The Best Accuracies are Presented in Bold.}}
\setlength\tabcolsep{6.5pt}
\renewcommand\arraystretch{1.1}
\begin{tabular}{c|c|ccc|cccc|c}
\hline
\multirow{2}{*}{\textbf{Class}} & \textbf{Conventional} & \multicolumn{3}{c|}{\textbf{CNN-based Methods}}  & \multicolumn{4}{c|}{\textbf{Transformer-based Methods}}     & \multirow{2}{*}{\textbf{3DSS-Mamba}} \\ \cline{2-9}
                                & SVM                   & 1D-CNN         & 2D-CNN         & 3D-CNN         & VIT(Pixel)     & VIT(Patch)     & SF             & HSI-BERT &                                      \\ \hline
1                               & 98.53                 & 90.25          & 98.40          & 97.04          & 93.34          & 91.56          & 99.29          & 92.26    & 99.02                                \\
2                               & 98.05                 & 95.14          & 99.20          & 98.48          & 99.29          & 94.69          & 98.05          & 83.15    & \textbf{99.56}                       \\
3                               & 98.88                 & \textbf{100.0} & \textbf{100.0} & \textbf{100.0} & \textbf{100.0} & \textbf{100.0} & \textbf{100.0} & 99.20    & \textbf{100.0}                       \\
4                               & 97.81                 & 93.89          & 98.31          & \textbf{98.95} & 87.41          & 98.48          & 96.43          & 93.47    & 97.68                                \\
5                               & 98.57                 & 93.40          & 99.28          & 99.44          & 99.37          & 92.75          & \textbf{100.0} & 99.73    & 99.91                                \\
6                               & 92.64                 & 99.38          & 92.62          & \textbf{100.0} & 97.26          & \textbf{100.0} & 95.89          & 78.76    & 98.29                                \\
7                               & 91.98                 & 80.84          & 94.01          & 93.38          & 85.28          & 97.11          & 96.06          & 91.06    & \textbf{97.55}                       \\
8                               & 91.96                 & 70.26          & 89.63          & 94.29          & 67.50          & 91.52          & 91.70          & 86.68    & \textbf{96.16}                       \\
9                               & 85.09                 & 67.65          & 83.87          & 93.61          & 70.63          & 87.67          & 94.59          & 89.60    & \textbf{96.27}                       \\
10                              & 93.16                 & 67.64          & 94.62          & 96.33          & 79.89          & 89.04          & 97.46          & 89.03    & \textbf{99.55}                       \\
11                              & 87.01                 & 69.64          & 90.53          & 96.84          & 58.63          & 93.88          & 90.56          & 95.31    & \textbf{98.20}                       \\
12                              & 87.25                 & 62.94          & 89.38          & 95.46          & 51.44          & 87.84          & \textbf{99.28} & 88.63    & 98.38                                \\
13                              & 32.35                 & 49.04          & 77.83          & 94.24          & 6.64           & 96.68          & 56.40          & 93.12    & \textbf{96.68}                       \\
14                              & 98.91                 & 97.20          & 99.30          & 98.60          & 79.48          & \textbf{99.74} & 98.44          & 97.92    & 99.48                                \\
15                              & 85.08                 & 99.09          & \textbf{100.0} & 99.85          & 98.65          & 99.66          & 99.66          & 99.66    & \textbf{100.0}                       \\ \hline
OA (\%)                         & 91.74                 & 81.06          & 93.93          & 96.77          & 79.27          & 93.63          & 95.46          & 91.67    & \textbf{98.37}                       \\
AA (\%)                         & 90.03                 & 82.42          & 93.80          & 97.10          & 78.32          & 94.70          & 94.25          & 91.84    & \textbf{98.44}                       \\
Kappa                           & 91.06                 & 79.51          & 93.44          & 96.51          & 77.54          & 93.12          & 95.09          & 90.99    & \textbf{98.24}                       \\ \hline
\end{tabular}
\label{tab:HU13}
\end{table*}

\subsubsection{Pavia University Dataset}
The classification experiments on the Pavia University dataset are conducted with 5\% of the reference samples. Table~\ref{tab:PU} provides the quantitative comparison results with each competitive approach. As can be observed, the proposed 3DSS-Mamba yields the most competitive performance in comparison with other studied methods. Restricted by hand-crafted feature descriptors, conventional approaches exhibit limitations in handling HSI data with complex contextual semantics, resulting in unsatisfactory classification performance. Despite achieving encouraging results, CNN-based models struggle to establish long-range dependencies due to their limited receptive fields. In contrast to Transformer-based architectures, the proposed 3DSS-Mamba performs global spectral-spatial contextual modeling from the sequence modeling perspective, achieving relatively stable and superior performance. Compared to the suboptimal approach, the quantitative improvements in terms of OA, AA, and Kappa up to 0.86\%, 0.81\%, and 1.14\%, respectively.
\begin{table*}[!htb]
\centering
\caption{\textsc{Computational parameter analysis of various comparison approaches in terms of parameters, Flops, and inference time on the PU dataset.}}
\setlength\tabcolsep{7.5pt}
\renewcommand\arraystretch{1.1}
\begin{tabular}{c|cccccccc}
\hline
\textbf{Metrics}                      & 1D-CNN & 2D-CNN & 3D-CNN & VIT(Pixel) & VIT(Patch) & SF     & HSI-BERT & 3DSS-Mamba \\ \hline
Params (M)                             & 0.0083 & 0.0101 & 0.0073 & 0.1107     & 0.1107     & 0.1222 & 0.3006   & 0.0103     \\
Flops (G)                              & 0.0015 & 0.0125 & 0.080   & 0.0139     & 1.1902     & 0.3292 & 5.4410    & 0.8936     \\
\multicolumn{1}{l|}{Inference time (s)} & 0.91   & 0.72   & 1.13   & 1.64       & 3.49       & 6.55   & 37.46    & 4.67       \\ \hline
\end{tabular}
\label{tab:Complex}
\end{table*}

\begin{figure*}[htbp]
	\centering
	\begin{subfigure}{0.32\linewidth}
		\centering
		\includegraphics[width=6cm,height=4cm]{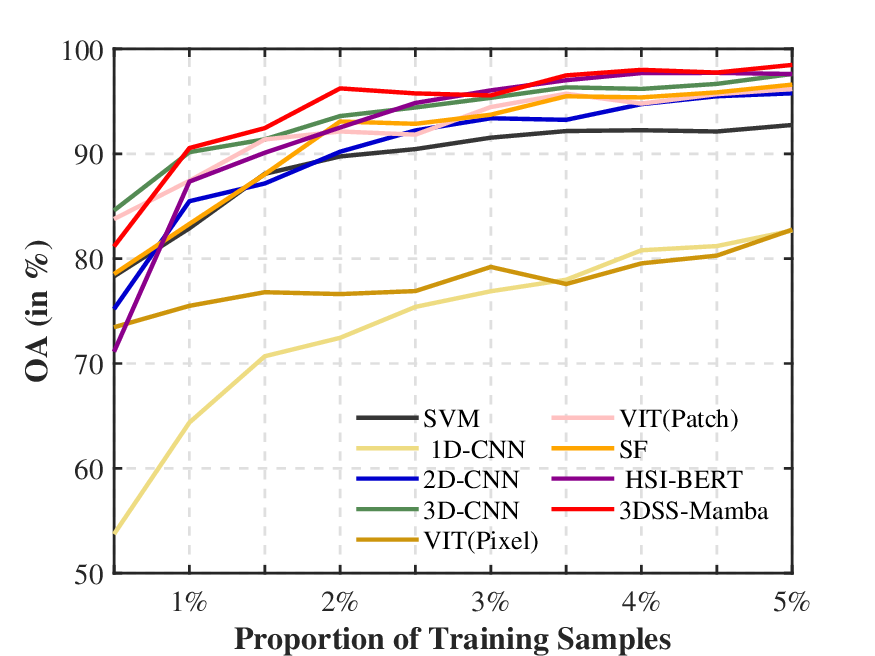}
		\caption{}
        \setlength{\belowdisplayskip}{43pt}
		\label{pseudo-I}
	\end{subfigure}
	\centering
	\begin{subfigure}{0.32\linewidth}
		\centering
		\includegraphics[width=6cm,height=4cm]{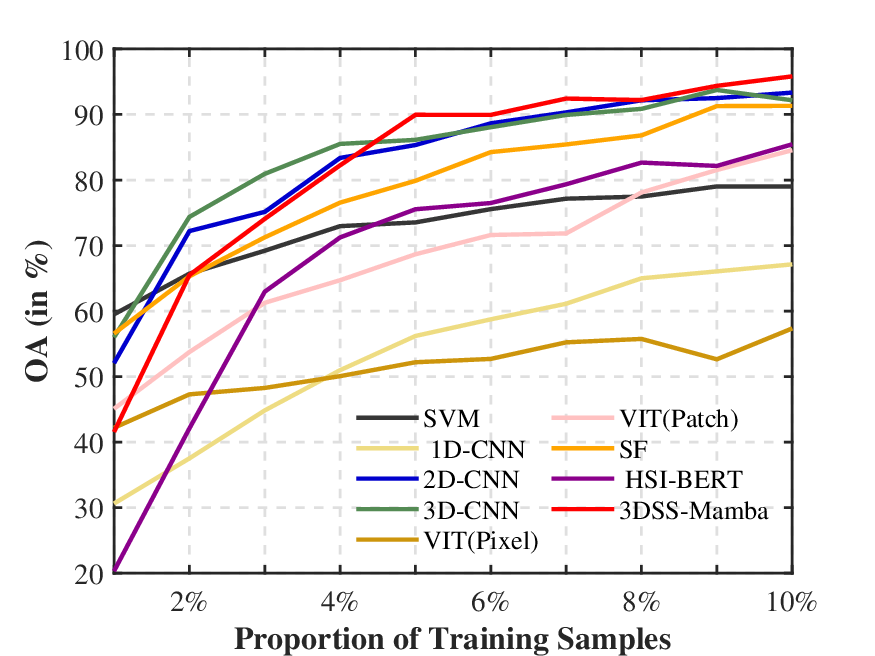}
		\caption{}
		\label{pseudo-I}
	\end{subfigure}
    \begin{subfigure}{0.3\linewidth}
		\centering
		\includegraphics[width=6cm,height=4cm]{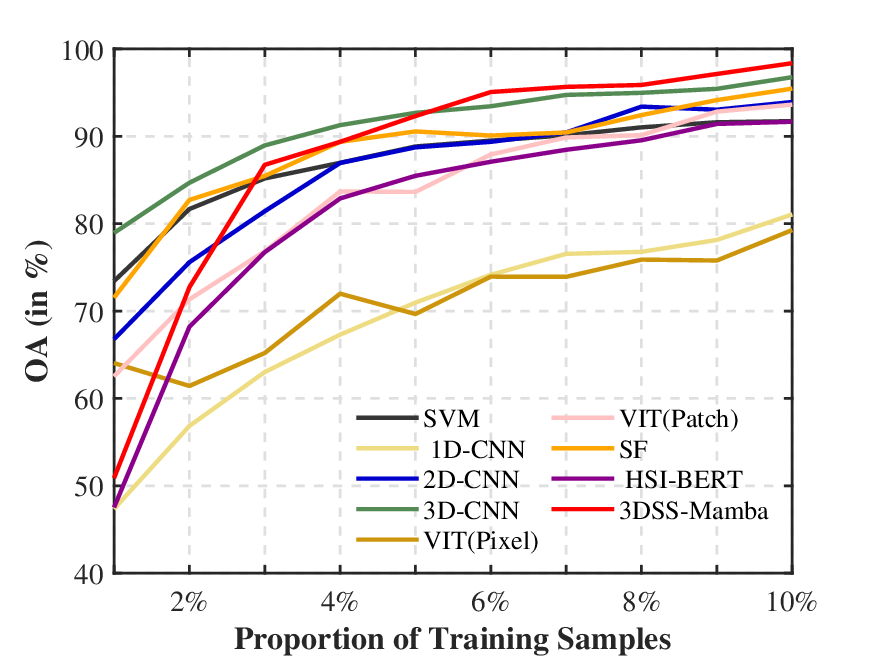}
		\caption{}
		\label{pseudo-I}
	\end{subfigure}
    \caption{Classification performance of the compared methods with different numbers of labeled data on three datasets. (a) Pavia University. (b) Indian Pines. (c) Houston 2013.}
	\label{fig:Ratio}
\end{figure*}

Additionally, the corresponding classification maps obtained by different approaches are visualized in Fig.~\ref{fig:PU}. As can be observed, conventional architecture such as SVM generally brings noticeable noises, which can be attributed to the limited feature extraction capability of hand-crafted descriptors. Typical CNN-based and Transformer-based approaches such as 1D-CNN and VIT(Pixel) suffer from significant misclassifications, particularly evident in the bare soil category. In contrast, the proposed 3DSS-Mamba achieves the most consistent results with the ground truth, which presents the clearest category boundaries with minimal noise.

\subsubsection{Indian Pines Dataset}
The experiments on the Indian Pines dataset are performed with 10\% of the reference samples. The quantitative classification accuracies are reported in Table~\ref{tab:IP}. Based on the results, the proposed 3DSS-Mamba achieves the highest recognition performance in comparison with other competitive approaches, and exhibits excellent improvements in several categories, such as Corn-notill, Corn-mintil, and Soybean-notill. Due to the lack of consideration of spatial information, the Transformer-based VIT(Pixel) method inevitably suffers from undesirable classification consequences, with the reduction compared to 3DSS-Mamba reaching 38.47\% for OA, 49.18\% for AA, and 46.24\% for Kappa, respectively. To highlight the differences in classification results, Fig~\ref{fig:IP} further provides the visualization maps of various methods. Benefiting from the extraction of global spectral-spatial semantic information with the 3D sequential modeling mechanism, the proposed 3DSS-Mamba exhibits the smoothest and clearest classifications across most regions despite slight edge information confusion. These phenomena further reveal the effectiveness and preeminence of the proposed method.

\subsubsection{Houston 2013 Dataset}
The experiments on the Houston 2013 dataset are executed with 10\% of the labeled samples. As evident from the classification results in Table~\ref{tab:HU13}, the proposed 3DSS-Mamba consistently outperforms other techniques by substantial margins, demonstrating the highest quantities across all three metrics. In contrast to the suboptimal method, 3DSS-Mamba achieves significant improvements in OA, AA, and Kappa by 1.6\%, 1.34\%, and 1.73\%, respectively. Regarding the visualization maps in Fig~\ref{fig:HU13}, 3DSS-Mamba provides the most precise prediction details, even though this scenario is predominantly distributed with discrete and local sample targets. These excellent improvements further verifies the potential application of sequence scanning model in HSI classification.

\begin{figure}[tb]
    \centering
	\includegraphics[width=8cm,height=6cm]{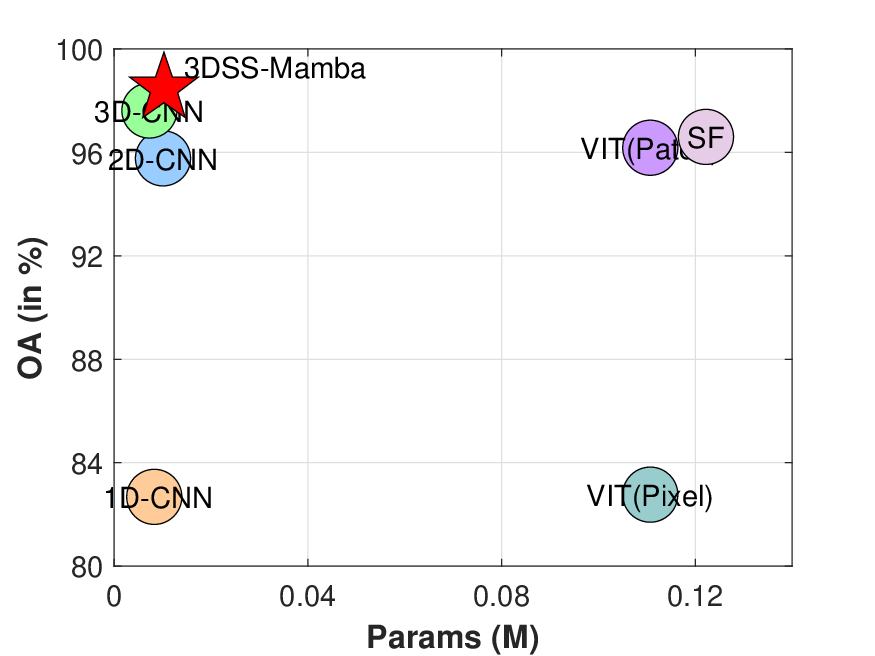}
	\caption{Computational parameter analysis of various comparison approaches on the PU dataset.}
\label{fig:Param}
\end{figure}

\subsection{Analysis of Computational Complexity}
This section investigates the computational complexity of the proposed 3DSS-Mamba, focusing on model parameters, Flops, and inference time. Fig~\ref{fig:Param} illustrates the model parameter sizes of various comparison approaches on the Pavia University dataset. Attributed to the inherent linear sequential modeling mechanism, the proposed 3DSS-Mamba achieves the best classification performance with significantly fewer computational parameters. Although the CNN-based methods enjoy slight computational burdens, their capability of capturing long-range dependencies is constrained by their local receptive field, restricting further performance improvement. Besides, Transformer-based methods generally suffer from higher resource consumption due to the series of multi-head self-attention (MHSA) modules. While VIT(Patch) and SF methods provide competitive performance, their model parameters are almost 10 times that of 3DSS-Mamba. Table~\ref{tab:Complex} further presents the detailed model parameters, Flops, and inference time. In summary, the proposed 3DSS-Mamba exhibits competitive advantages in balancing computational efficiency and classification effectiveness, highlighting significant potentiality and viability for HSI classification tasks.

\subsection{Robustness Assessment}
To demonstrate the robustness of the proposed 3DSS-Mamba, extensive experiments are conducted considering various proportions of training samples. Specifically, the selected percentage for the Indian Pines and Houston 2013 datasets covers an interval of $\left\{ {1.0\% ,2.0\% , \ldots ,10.0\% } \right\}$, and for the Pavia University is set to $\left\{ {0.5\% ,1.0\% , \ldots ,5.0\% } \right\}$. Fig~\ref{fig:Ratio} displays the performance variations with different percentages on the four HSI datasets, with the proposed 3DSS-Mamba highlighted by red curves. There is a basically reasonable trend that the classification accuracy of 3DSS-Mamba steadily increases with the percentages of training samples, which exhibit substantial robustness. Furthermore, 3DSS-Mamba consistently outperforms other competitive methods across most percentages on the three testing datasets. These notable advantages further demonstrate the feasibility and superiority of the proposed method.

\section{Conclusion}
\label{sec:conc}
In this paper, we introduce 3D-Spectral-Spatial Mamba (3DSS-Mamba), a novel architecture based on the State Space Model (SSM) for HSI classification. Benefiting from the integrated Spectral-Spatial Token Generation module (SSTG) and 3D-Spectral-Spatial Selective Scanning (3DSS) mechanism, 3DSS-Mamba achieves the substantial advantages of both global spectral-spatial contextual modeling and linear computational complexity from the sequence modeling perspective. Extensive experiments demonstrate that the proposed 3DSS-Mamba efficiently breaks the performance and efficiency bottlenecks of state-of-the-art CNN-based and Transformer-based HSI architectures. This research offers a feasible solution for the HSI classification task. Future work will endeavor to explore the scalability of the Mamba model across a wider range of hyperspectral scenarios.

\bibliographystyle{IEEEtran_doi}
\bibliography{deepcs}
\renewenvironment{IEEEbiography}[1] {\IEEEbiographynophoto{#1}}{\endIEEEbiographynophoto}

\end{document}